\documentclass[conference]{IEEEtran}
\IEEEoverridecommandlockouts

\usepackage{cite}
\usepackage{amsmath,amssymb,amsfonts}
\usepackage{algorithmic}
\usepackage{graphicx}
\usepackage{textcomp}
\usepackage{xcolor}
\usepackage{hyperref}
\usepackage{amssymb}
\usepackage{amsmath}
\usepackage{makecell}
\usepackage{adjustbox}
\usepackage{comment}

\usepackage[table]{xcolor}
\definecolor{graycolour}{RGB}{192,192,192}

\newcommand{\myeq}{\mathrel{\overset{\Delta}{=}}}
\newcommand{\argmin}{\arg\!\min}

\def\BibTeX{{\rm B\kern-.05em{\sc i\kern-.025em b}\kern-.08em
    T\kern-.1667em\lower.7ex\hbox{E}\kern-.125emX}}
\begin{document}

\title{XDoGE: Multilingual Data Reweighting to Enhance Language Inclusivity in LLMs\\
\thanks{\IEEEauthorrefmark{2} Equal contribution.}
}

\author{\IEEEauthorblockN{Iñaki Lacunza\IEEEauthorrefmark{2}}
\IEEEauthorblockA{\textit{Language Technologies Lab} \\
\textit{Barcelona Supercomputing Center}\\
Barcelona, Spain \\
inaki.lacunza@bsc.es}
\and
\IEEEauthorblockN{José Javier Saiz\IEEEauthorrefmark{2}}
\IEEEauthorblockA{\textit{Language Technologies Lab} \\
\textit{Barcelona Supercomputing Center}\\
Barcelona, Spain \\
jose.saiz@bsc.es}
\and
\IEEEauthorblockN{Alexander Shvets\IEEEauthorrefmark{2}}
\IEEEauthorblockA{\textit{Language Technologies Lab} \\
\textit{Barcelona Supercomputing Center}\\
Barcelona, Spain \\
aleksandr.shvets@bsc.es}
\and
\IEEEauthorblockN{Aitor Gonzalez-Agirre
}\IEEEauthorblockA{\textit{Language Technologies Lab} \\
\textit{Barcelona Supercomputing Center}\\
Barcelona, Spain \\
aitor.gonzalez@bsc.es}
\and
\IEEEauthorblockN{Marta Villegas}
\IEEEauthorblockA{\textit{Language Technologies Lab} \\
\textit{Barcelona Supercomputing Center}\\
Barcelona, Spain \\
0000-0003-0711-0029}
}

\maketitle

\begin{abstract}
Current large language models (LLMs) are trained on massive amounts of text data, primarily from a few dominant languages. Studies suggest that this over-reliance on high-resource languages, such as English, hampers LLM performance in mid- and low-resource languages. To mitigate this problem, we propose to (i) optimize the language distribution by training a small proxy model within a domain-reweighing DoGE algorithm that we extend to XDoGE for a multilingual setup, and (ii) rescale the data and train a full-size model with the established language weights either from scratch or within a continual pre-training phase (CPT). We target six languages possessing a variety of geographic and intra- and inter-language-family relations, namely, English and Spanish (high-resource), Portuguese and Catalan (mid-resource), Galician and Basque (low-resource). We experiment with Salamandra-2b, which is a promising model for these languages. We investigate the effects of substantial data repetition on minor languages and under-sampling on dominant languages using the IberoBench framework for quantitative evaluation. Finally, we release a new promising IberianLLM-7B-Instruct model centering on Iberian languages and English that we pretrained from scratch and further improved using CPT with the XDoGE weights.
\end{abstract}


\section{Introduction}
The development of large language models (LLMs) has predominantly focused its advances on high-resource languages like English, Spanish or Chinese \cite{Grattafiori, Adler, Yang, Young}.
Given that most other languages may not have enough data to train LLMs from scratch, the only option left for them is to be included among the target languages of a multilingual model. Multilingual LLMs have the potential to handle multiple languages comprehensively; however, approaches to dealing with unbalanced pre-training data have traditionally been conservative and have not specifically addressed how to improve performance in underrepresented languages, leaving their contribution to the mixture of training data residual compared to dominant languages \cite{Scao, Jiang, Martins}.

\begin{figure}[t!]
    \centering
    \includegraphics[width=\columnwidth]{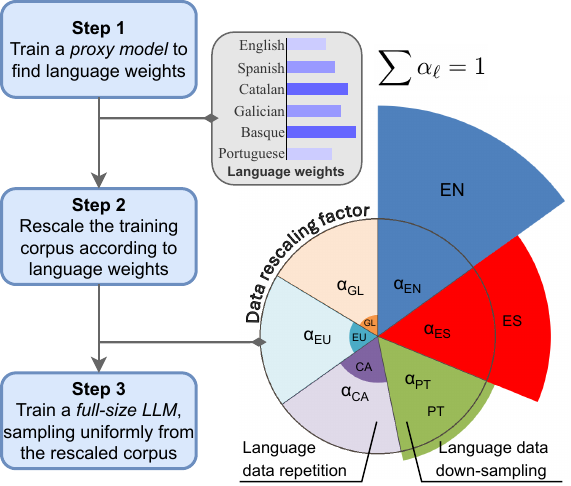}
    \caption{Summary of our proposed XDoGE method.}
    \label{fig:xdoge_summary}
\end{figure}

We propose to enhance the linguistic diversity of LLMs by building upon a state-of-the-art method in domain reweighting, DoGE \cite{Fan}, which optimizes the distribution of sources, distinguishing those that help the most to model a language within a given set of domains. In a pilot study, \cite{Fan} showed the potential of their approach (primarily designed for domains in a single language) to generalize from a mixture of high-resource languages to a certain out-of-mixture target language, when there is a grammatical relatedness. However, the possibility of generalization to several languages at once as well as addressing constraints on data availability for resampling the pre-train corpus remained unexplored.

We adapt the DoGE method to a multilingual scenario, resulting in XDoGE. In particular, we modify DoGE to handle multilingual sources that we split beforehand by language and weight individually, thereby avoiding reliance on a prior per-source language distribution, which is often skewed towards high-resource languages. We also introduce a weight-clipping mechanism so that no language becomes excessively over- or under-represented to promote fair modeling of all the target languages. We also rescale the training corpus to satisfy the optimized language distribution. This allows us to down-sample data for dominant languages and compensate for the lack of data in mid- and low-resource languages without compromising the model's performance for most languages. The summary of our approach is provided in Fig.~\ref{fig:xdoge_summary}.

We show that XDoGE succeeds in generalizing to languages with distinct grammars, and overcomes the uniform-weighting baseline in many languages when small base models are trained from scratch as well as within a continual pre-training phase (CPT) for larger models.

Our main contributions are:
\begin{itemize}
    \item We introduce the XDoGE framework, based on DoGE, which re-weights sources in multilingual pre-training data for improved generalization on the target languages. We add mechanisms to handle multiple sources in different languages in an imbalanced data scenario (see Section~\ref{sec:methodology}).
    \item We demonstrate that XDoGE is scalable to various proxy and base model sizes and yields improvements for many languages in both from-scratch pre-training and continual pre-training scenarios, as measured by perplexity and a range of downstream metrics in few-shot tasks (see Section~\ref{sec:experiments}). Note that in studies with relatively small models within a limited token budget, we could aim only at minimal signals of linguistic proficiency improvement.
    \item We also release IberianLLM-7B-Instruct\footnote{\url{https://huggingface.co/langtech-languagemodeling/IberianLLM-7B-Instruct}}\footnote{We release the instruction-tuned version of the model developed in this work to facilitate its integration into general-purpose applications. See the model card for details on instruction-tuning.} centered in Iberian languages and English, well-performing in various high-, medium-, and low-resource languages that meet our aim to promote linguistic inclusion in the field.
\end{itemize}

\section{Related Work}

While there have been significant efforts to develop multilingual LLMs, most attempts have sampled pre-training data with heuristic approaches, which lack generalization and still may be biased towards high-resource languages. In most cases, heuristic approaches only consider data size and ignore language similarity, which has been found important for cross-lingual transfer learning \cite{Chung, Nigatu, Gogoulou}.

On the other hand, it is a natural assumption in machine learning that principled techniques yield more generalizable weightings than manually tuned hyperparameters \cite{Lee}. Among the latest principled techniques, ``offline'' data weighting algorithms optimize the domain weights before actually training the language model, providing an optimized distribution of data from the start. DoReMi \cite{Xie} optimizes domain weights by aiming to achieve low loss on all domains, formalizing this problem as a group distributionally robust optimization \cite{Oren, Sagawa}. Specifically, DoReMi optimizes the domain weights with a proxy model by reducing the excess loss on each domain compared to the loss on a previous reference model, and then uses the optimized domain weights to train a larger model. More recently, the DoGE algorithm \cite{Fan} has been proposed and shown to be scalable to different model sizes. Similar to DoReMi, the algorithm involves a two-step process by first training a small \textbf{proxy} model that uses the gradient alignment between each domain and all other target domains in order to give more weight to domains with better transfer to other domains. Second, a larger \textbf{base} model is trained using data sampled according to the final weight distribution. However, to the best of our knowledge, there has not yet been an attempt to reproduce these principled approaches in a setting that covers many target languages. There are studies on multilingual data mixing, but the weights are often chosen manually and concern data types rather than languages \cite{Ustun}.

\section{Methodology}
\label{sec:methodology}

Building on the offline domain reweighting algorithm DoGE \cite{Fan}, we adapt it to enhance cross-lingual transfer in model pre-training for a set of target languages, resulting in XDoGE. Our adaptation concerns (i) the ability to target a combination of languages and domains within a multilingual corpus aiming to weight language-domain sources based on their contribution to the overall cross-lingual learning; (ii) an introduction of a threshold mechanism that ensures an adequate number of instances for each language used at every distribution update step, maintaining its representation up to the weight convergence; (iii) a corpus rescale assessment to balance an undesired cut in data for high-resource languages and excessive data repetition for low-resource languages. In the following, we outline XDoGE, adapting the original notation.\footnote{We disclose only core elements of DoGE. Cf. \cite{Fan}, for the individual decisions in the optimization design.}

\subsection{Proxy Model Training.}
Let $S$ denote the set of sources (domains) and $\mathcal{L}$ -- the set of languages with $k_\ell$ sources in a language $\ell \in \mathcal{L}$: $S_\ell = \{S_{\ell,1}, S_{\ell,2}, \dots, S_{\ell,k_\ell}\}$ (e.g., $S_{ES,1}$ -- Spanish Wikipedia,\footnote{\href{https://es.wikipedia.org/}{https://es.wikipedia.org/}} $S_{ES,2}$ -- Spanish part of Community OSCAR\footnote{Subsets with the \texttt{es} code from \url{https://huggingface.co/datasets/oscar-corpus/community-oscar}}). We optimize sampling weights for all sources $k = \sum_{\ell \in \mathcal{L}} k_\ell$ over the
probability simplex $\alpha \in \Delta_k \subset \mathbb{R}^k$ using an original bi-level optimization framework adjusted to our split of domains into language-based subsets with an addition of a minimum weight constraint:\vspace{-3mm}

    \[\alpha \in \argmin_{\alpha \in \Delta_k} \sum_{\ell \in \mathcal{L}} \sum_{i=1}^{k_\ell} l_{\ell,i}(\theta^{*}(\alpha)), \]\vspace{-2mm}
    \[s.t. \ \ \theta^{*}(\alpha) \in \argmin_{\theta} \sum_{\ell \in \mathcal{L}} \sum_{i=1}^{k_\ell} \alpha_{\ell,i} l_{\ell,i}(\theta), \]  
    \[s.t. \ \ \alpha_{min} \geq \gamma; \gamma \lll |S|^{-1},\]
where  $l_{\ell,i}(\theta)$ is the next-token prediction (cross-entropy) loss of the model of parameters $\theta$ on $S_{\ell,i}$, and $\gamma$ is a constant.

A train batch at time-step $t$ is sampled from the dynamically updated instance-wise distribution:
\begin{equation}
  P_{\alpha} \myeq
    \sum_{\ell\in\mathcal{L}} \sum_{i=1}^{k_\ell}
      \alpha^{(t)}_{\ell,i}\, \cdot \mathtt{UNIF}\!\big(S_{\ell,i}\big).
\end{equation}

where $\alpha_{\ell,i}$ is the weight for source $S_{\ell,i}$, and $\texttt{UNIF}(S_{\ell,i})$ is a uniform batch-sampling over $S_{\ell,i}$.

\subsection{Bi-level Optimization.}
The framework alternates between two key steps: the inner loop (\ref{eq:inner}), which updates the model parameters $\theta$ using weighted losses, and the outer loop (\ref{eq:outer1}-\ref{eq:outer4}) which updates weights $\alpha$ based on gradient alignments across sources (domain-language pairs that are better aligned with the other pairs receive a larger weight).

The $\theta$ parameters are updated as (\textit{the inner loop}):
\begin{equation}
\label{eq:inner}
    \theta^{(t+1)} = \theta^{(t)} - \eta^{(t)} \sum_{\ell \in \mathcal{L}} \sum_{i=1}^{k_\ell} \alpha_{\ell,i}^{(t)} \nabla l_{\ell,i}(\theta^{(t)}),
\end{equation}
where $\alpha^{(t)}$ is used to re-weight the loss from sources at time-step $t$, $\eta(t)$ is the step size, and $\nabla l_{\ell,i}(\theta(t))$ is a stochastic gradient for $S_{\ell,i}$ samples.

The weights $\alpha$ are updated as (\textit{the outer loop}):

\begin{equation}
\label{eq:outer1}
    W_{\ell,i}^{(t)} = \left\langle \nabla l_{\ell,i}(\theta^{(t)}), \sum_{\ell' \in \mathcal{L}} \sum_{j=1}^{k_{\ell'}} \nabla l_{\ell',j}(\theta^{(t)}) \right\rangle,
\end{equation}
\begin{equation}
\label{eq:outer2}
    \hat{\alpha}_{\ell,i}^{(t)} = \alpha_{\ell,i}^{(t-1)} \odot \exp\left(\frac{\eta^{(t)} W_{\ell,i}^{(t)}}{\mu}\right),
\end{equation}
\begin{equation}
\label{eq:outer3}
   \tilde{\alpha}_{\ell,i}^{(t)} = \frac{\hat{\alpha}_{\ell,i}^{(t)}}{\sum_{\ell' \in \mathcal{L}} \sum_{j=1}^{k_{\ell'}} \hat{\alpha}_{\ell',j}^{(t)}},
\end{equation}
\begin{equation}
\label{eq:outer4}
    \alpha_{\ell,i}^{(t)} = \texttt{Proj}_{\Delta_{k,\gamma}}\left(\tilde{\alpha}_{\ell,i}^{(t)}\right).
\end{equation}
where $W_{\ell,i}^{(t)}$ is the stochastic generalization estimation function (a higher value means learning $S_{\ell,i}$
will also contribute to learning other languages and domains), $\mu$ - a hyperparameter for the strength of regularization, $\texttt{Proj}_{\Delta_{k,\gamma}}$ -- our new projection operator that ensures $\alpha_{\ell,i} \geq \gamma$ $\forall \ell,i$ while preserving $\sum_{\ell,i} \alpha_{\ell,i} = 1$. The projection is applied as:
\begin{enumerate}
    \item Clip weights: $\alpha'_{\ell,i} = \max(\tilde{\alpha}_{\ell,i}, \gamma)$
    \item Compute excess weight: $E = \sum_{\ell,i} \alpha'_{\ell,i} - 1$
    \item Redistribute $E$ proportionally from unclipped sources:
\end{enumerate}
\begin{equation}
    \alpha_{\ell,i}^{(t)} = 
    \begin{cases}
        \alpha'_{\ell,i} - E\cdot \frac{ \alpha'_{\ell,i}}{\sum_{(\ell',j) \in U} \alpha'_{\ell',j}}, & (\ell,i) \in U \\
        \gamma, & \text{otherwise},
    \end{cases}
\end{equation}
where $U = \{(\ell,j) \mid \alpha'_{\ell,j} > \gamma\}$. This prevents language under-representation while promoting domain diversity within a language. This threshold mechanism adds negligible overhead compared to standard training, preserving DoGE's computational efficiency.

\subsection{Final Language Weight Assignment.}
As the algorithm avoids relying on the prior domain size and therefore the obtained weights within a language could be excessively disproportional to the number of tokens per domain available in the corpus for language modeling, we assign a single domain-independent language weight for the final sampling. The final language weight \(\bar{\alpha_{\ell}} \) is computed as the sum of its domain weights:\vspace{-3mm}

\begin{equation}
    \bar{\alpha_{\ell}} = \sum_{i=1}^{k_\ell} \alpha_{\ell,i}.
\end{equation}
Finally, training instances are sampled for each batch based on the language weights:

\begin{equation}
  P_{\bar{\alpha}} \myeq
    \sum_{\ell\in\mathcal{L}} \bar{\alpha_{\ell}}\;\cdot\operatorname{UNIF}\!\big(S_{\ell}\big).
\end{equation}
This mitigates the lack of domain-language data to support the allocated weight and makes large-scale training practically feasible.

\subsection{Corpus Rescaling at Large Model Training.}

The data for full-size model training is usually highly unbalanced across languages. Therefore, if a language is assigned a higher probability than its token count in a corpus would suggest, the corresponding part of the corpus will be sampled several times. Conversely, if the assigned weight is lower, the corresponding language instances will be sampled fewer times than the corpus allows (see Fig.~\ref{fig:xdoge_summary}).

In our approach, we aim to find a trade-off between over- and under-sampling by continuously evaluating intermediate checkpoints, resulting in a data rescaling factor respectful of all languages. This ensures that low-resource languages, gaining more repetitions, are sufficiently represented without causing model degradation, while for high-resource languages, a considerable portion of the available data is utilized. We train the \textit{base} models from scratch, and \textit{pre-trained} models within a continual pre-training phase (CPT). In both cases, we use the standard next-token prediction loss.

\section{Experiment Setup}

\subsection{Data}
We collect a training dataset to test the effect of languages on cross-lingual transfer during pre-training and continual pre-training. We choose a set of 6 languages, including English, Spanish, Catalan, Galician, Basque and Portuguese, which represent a group of four typologically related Romance languages (along with English, which is typologically more distant), with varying degrees of availability of NLP resources \cite{Joshi}. Basque, on the other hand, is a language isolate that provides a contrast to the Romance languages, which can help to assess how structural and lexical similarities, as well as resource availability, influence cross-lingual transfer performance. We note that our approach can be readily extended to other languages beyond those selected in this study.

For the XDoGE base and proxy trainings, we source the data from Wikipedia dumps from May 2024, as it consists of roughly similar encyclopedic register across languages; and from Community OSCAR\footnote{We used 35 monthly dumps of the Community OSCAR corpus.}
\cite{Brack}, which is a processed multilingual corpus from web-crawled data, and represents a more heterogeneous source of data for each language. Both sources are widely used for the training of LLMs \cite{Scao}. For the CPT phase, we replace web data with FineWeb-Edu\footnote{\href{https://huggingface.co/datasets/HuggingFaceFW/fineweb-edu}{datasets/HuggingFaceFW/fineweb-edu}} for English and FineWeb2\footnote{\href{https://huggingface.co/datasets/HuggingFaceFW/fineweb-2}{datasets/HuggingFaceFW/fineweb-2}} for other languages \cite{Penedo} to ensure that pre-trained models that used OSCAR as a major source are exposed to new information, allowing them to show improvement with a smaller token budget.

Within each dataset, we perform exact document deduplication for all languages (to control the information repetition across training epochs better) and random shuffling followed by tokenizing with the six-language tokenizer described in Section~\ref{subsec: technical setup}. The dataset sizes in tokens are shown in Table~\ref{tab:data}.

\begin{table}[!ht]
    \centering
    \caption{Size in billions (B) of tokens of the training corpus for each language, where proxy and base trainings used OSCAR and Wikipedia, and CPT trainings used FineWeb-Edu, FineWeb2, and Wikipedia. Validation and test sets represent each 1\% of the training data.}
    \resizebox{\columnwidth}{!}{
        \begin{tabular}{c|ccc}
        \hline
            Language & OSCAR & Wikipedia & FineWeb2/-Edu \\ \hline
            English & 327,98 B & 4,76 B & 198,54 B \\ 
            Spanish & 140,61 B & 1,19 B & 159,36 B \\ 
            Portuguese & 62,12 B & 0,59 B & 117,00 B \\
            Catalan & 3,48 B & 0,45 B & 9,37 B \\ 
            Basque & 0,33 B & 0,12 B & 3,29 B \\             Galician & 0,11 B & 0,10 B & 1,50 B \\ \hline
            Total & 534,62 B & 7,21 B & 489,07 B \\ \hline
        \end{tabular}
    }
  \label{tab:data}
\end{table}

\subsection{Technical Setup}
\label{subsec: technical setup}

\paragraph{Proxy models}
While \cite{Fan} run the experiments at three different scales up to 125M parameters and demonstrate effective weight estimation at \(10k\) training steps within a monolingual setup, we allocate more resources to account for potentially more demanding domain-language interactions and consider proxy models with 70M, 125M, 250M, and 500M parameters (see Table~\ref{tab:proxy_dimensions}).\footnote{Rounded sizes correspond to actual of \(68M\), \(122M\), \(273M\), and \(509M\).} We perform longer runs of \(10k\)-\(50k\) steps to validate the robustness of our approach across different-sized models.

\begin{table*}
  \centering
  \caption{\label{tab:proxy_dimensions} Dimensions of the proxy models. The LLaMA-2 architecture has been used to build these models.}
  \begin{tabular}{lllll}
    \hline
    \textbf{Model ID} & \textbf{Proxy-70M} & \textbf{Proxy-125M} & \textbf{Proxy-250M} & \textbf{Proxy-500M} \\
    \hline
    Parameters & 67,936,768 & 122,358,528 & 273,122,304 & 509,199,360 \\
    Layers & 8 & 12 & 16 & 24 \\
    Hidden Size & 512 & 768 & 1,024 & 1,024 \\
    FFN Size & 512 & 512 & 2024 & 4096 \\
    Attention Heads & 8 & 12 & 16 & 16 \\
    K/V Heads & 8 & 12  & 16 & 16 \\
    Context Length & 128 & 128 & 128 & 128 \\
    Vocabulary Size & 52,000 & 52,000 & 52,000 & 52,000 \\
    \hline
  \end{tabular}
  
\end{table*}

To train the proxy models, we use the DoGE PyTorch-based framework\footnote{\href{https://github.com/Olivia-fsm/doge}{https://github.com/Olivia-fsm/doge}} and choose LLaMA architecture \cite{Touvron} to be compatible with Salamandra \cite{Gonzalez} used for the \textit{base} and \textit{pretrained models} in this work.

We use a maximum context length of 128 with a global batch size of 16,384 tokens, a learning rate of \(5e^{-4}\), a linear warmup cosine weight decay of \(1e^{-2}\), a warmup ratio of 0.05, and 500 warmup steps. We set $\gamma = 0.02$ empirically: with a batch size of 128 document instances, this threshold ensures that at least several instances from each source would be seen in each iteration.

\paragraph{Base models}
We train 250M, 500M and 900M base models with several reweighted data configurations (the resulting optimized weights are provided in Section~\ref{subsec:proxy_training}) and compare them in their respective sizes against \textit{baselines} that we train with uniform source weights, which is the preferred heuristic for cross-lingual generalization without prior knowledge on language relatedness.
Each model has a context length of \(8,192\) tokens and was trained with a global batch size of \(\sim4M\) tokens, equivalent to \(512\) instances per batch.

We train the models for about \(150k\) steps to observe up to 30\% of the English corpus, although making an extreme number of repetitions for downstream languages (e.g., up to 450 repetitions of Galician; cf.~Fig.~\ref{fig:xdoge_summary} for better illustration).

The base models were trained using the NeMo framework\footnote{\href{https://docs.nvidia.com/nemo-framework/}{https://docs.nvidia.com/nemo-framework/}} for its efficient parallelization and scalability in large-scale training. We use the architecture of Salamandra to ease the comparison with the \textit{pre-trained models} from the same model family described in the next subsection.

\paragraph{Pretrained models}
To mitigate excessive data repetition as happens with the base models, we tune already pre-trained models in a continual pertraining stage (CPT) that potentially requires fewer data to adapt to target languages (we limit to \(80k\) steps).

We selected the 2b-parameter variant from the Salamandra family of models \cite{Gonzalez}. This model was originally pre-trained on a corpus spanning 35 European languages and programming code, including all the languages targeted in our work.

We also pretrain IberianLLM-7b of the Salamandra architecture from scratch, focusing on six target languages (using the Salamandra pretraining data for corresponding languages).

For both models, we use Salamandra's tokenizer designed for multilingual support, based on byte-pair encoding \cite{Sennrich} with a vocabulary of 256,000 tokens and an equal number of documents per language to ensure fair representation across languages \cite{Dalt, Gonzalez}. We train the tokenizer on 6 languages and use it for proxy, base, and IberianLLM models. For Salamandra-2b we use its pre-trained tokenizer.

\section{Experiments}
\label{sec:experiments}

In this section, we discuss the findings from training the XDoGE proxies and its version without the thresholding mechanism (unthresholded proxy), as well as the choice of final language weights averaged from several runs and the results from applying XDoGE weights to from-scratch and continuous pre-training scenarios. We use a comprehensive set of few-shot tasks from a recently published benchmark for the Iberian languages, IberoBench \cite{Baucells}, to evaluate the models.

IberoBench comprises 60+ tasks that evaluate various language capabilities of LLMs. The tasks are of two main types: multiple choice (\textit{accuracy} metric)
and open-ended generation (BLEU \cite{Papineni} and ROUGE \cite{Lin}). It extends LM-Evaluation Harness,\footnote{\url{https://github.com/EleutherAI/lm-evaluation-harness}} reusing the Harness execution pipeline but supplies additional task YAMLs, dataset splits and prompt templates for Iberian languages. We run experiments with the same IberoBench and Harness configurations; therefore, the prompt text, the five examples (chosen with a fixed default seed for reproducibility), their order and formatting, and generation parameters are identical across models for all languages.

\subsection{Proxy Model Training}
\label{subsec:proxy_training}

\begin{figure}[ht]
\includegraphics[width=0.32\linewidth]{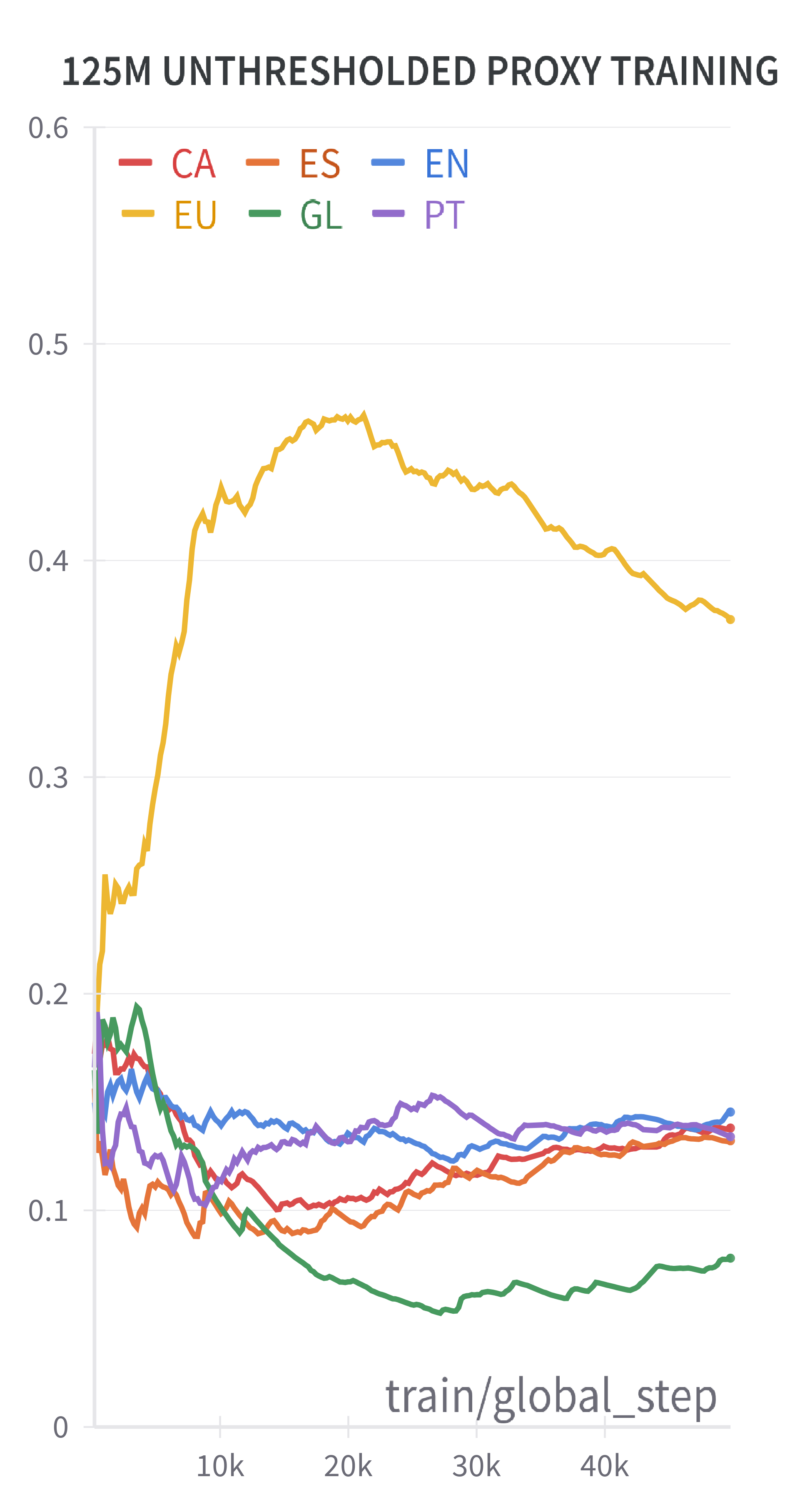}
\includegraphics[width=0.32\linewidth]{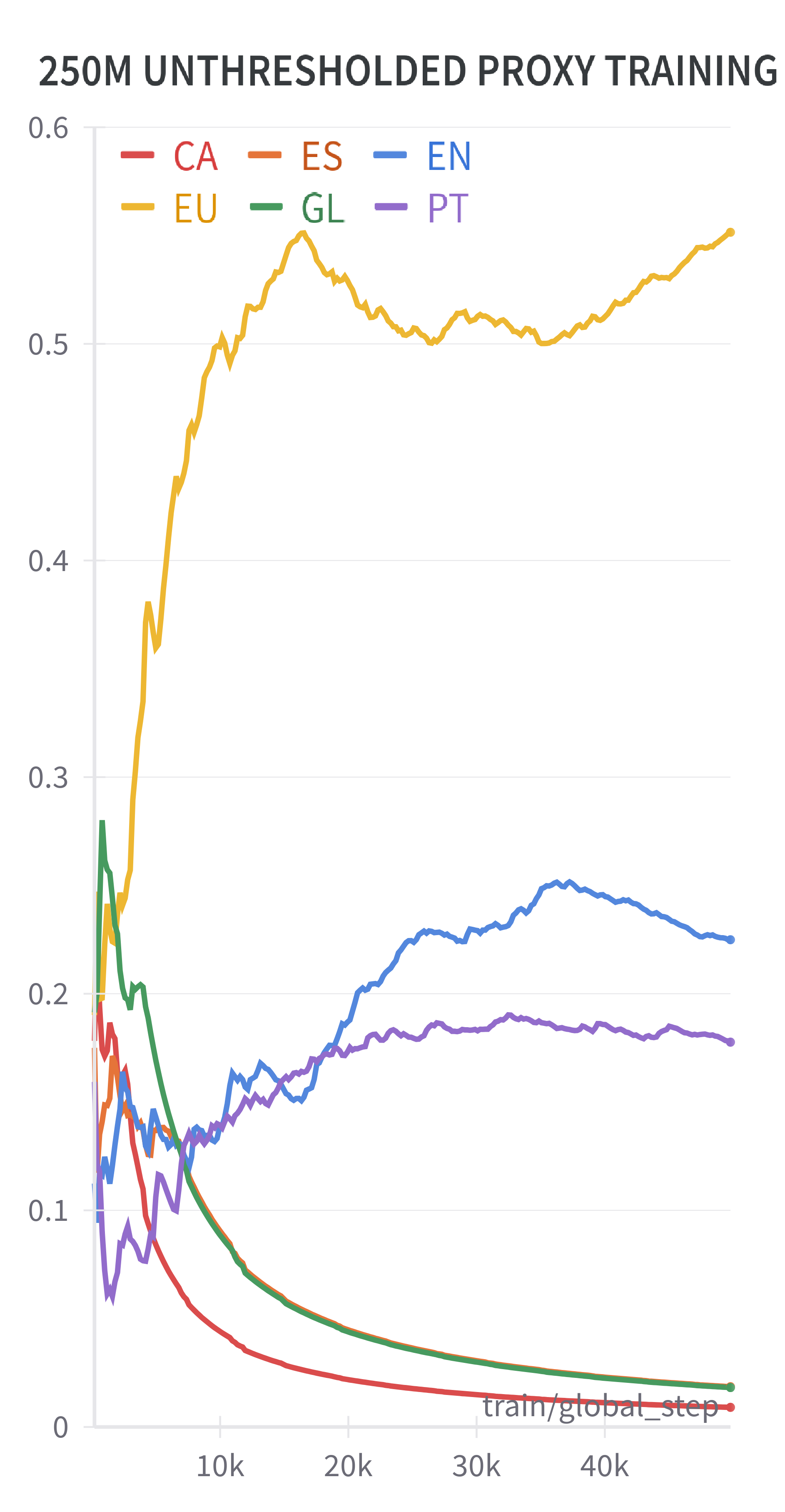}
\includegraphics[width=0.32\linewidth]{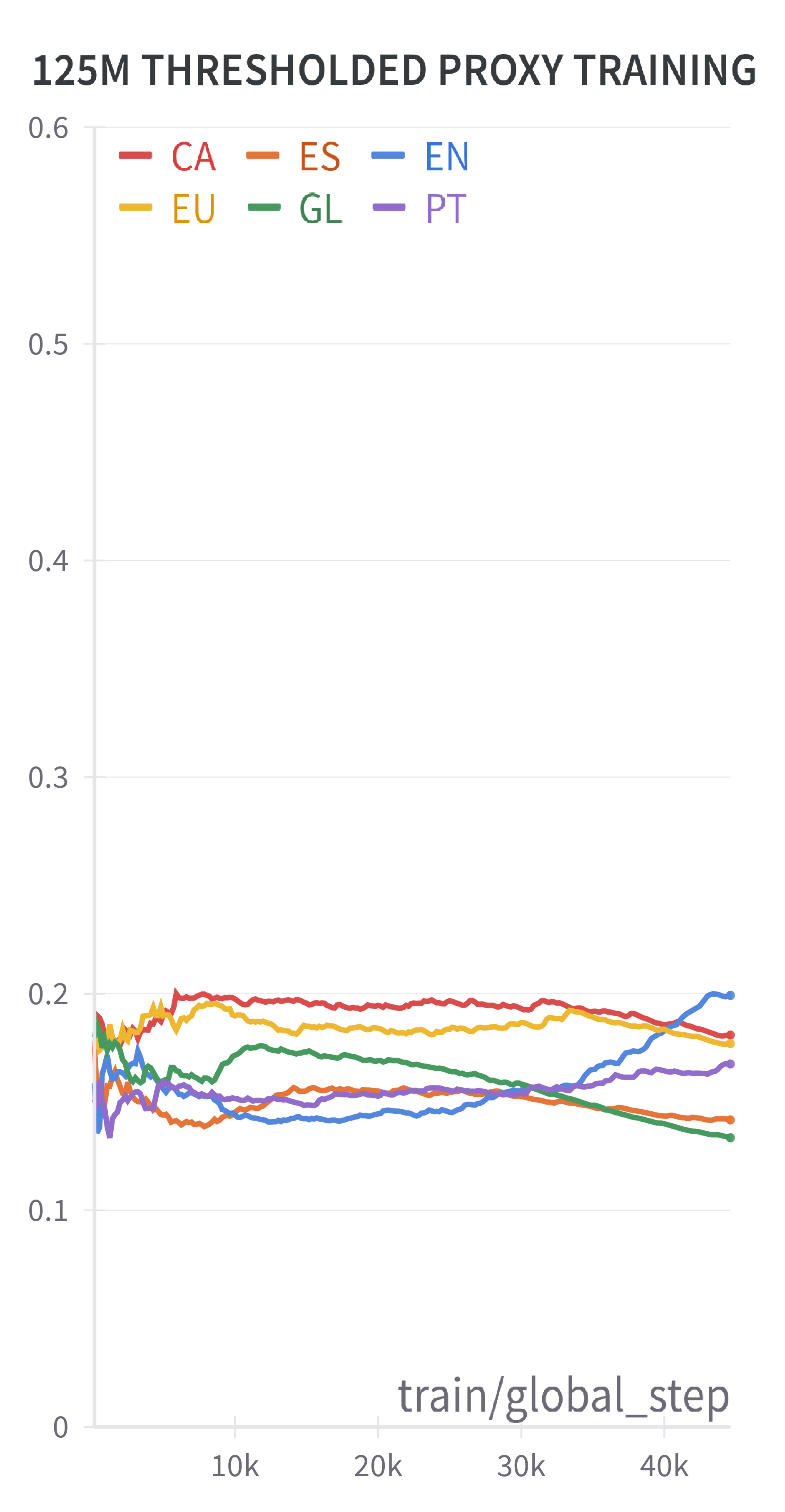}
    \caption{Language weight updates during the proxy model training without (125M and 250M on the left) and with the threshold mechanism (125M on the right).}
    \label{fig: unth vs th}
\end{figure}

Training unthreshold proxy models often results in some domain language weights dropping and stabilizing at near-zero values soon after the initial training steps (see Fig.~\ref{fig: unth vs th}). Since the proxy model is trained with the updated weights at each step, once a given language reaches near-zero weight values, the proxy model no longer receives a meaningful amount of data from that language and cannot recover its weight. On the other hand, languages that are more distinct, and therefore more ``difficult'' to learn, may receive increasingly larger weights if the threshold is not used (e.g., as seen with Basque). This may negatively affect the entire distribution.

Table~\ref{tab:threshold} shows the benefits of using a minimum threshold for domain-language weights: a) XDoGE converges stably to more similar distributions,\footnote{Compare $D_{KL}$ values for language weights.} avoiding drastic skews towards certain languages; b) the unthresholded variants often reverse the decision about which domain within a language to take for most of the tokens,\footnote{Compare $D_{KL}$ values for domain-language weights.} (both or only one, with strong but not exclusive preferences for Wikipedia of higher data quality), while the thresholded setup ensures that both domains contribute rather equally to the modeling of the language (no domain gets the minimum possible weight). This leads to the favorable conclusion that \textit{\textbf{learned weights can be assigned to languages independent of the domain}}, providing flexibility in domain selection and token assignment without having to retrain proxy models when new domains are added.

\begin{table*}[ht!]
  \centering
  \caption{Domain-Language (D-L) and Language (L) weights with different sizes of proxy models smoothed over 10k training steps. $D_{KL}$ -- KL divergence $\cdot 100\%$; against the distribution obtained by the 500M models.}
  \resizebox{\textwidth}{!}{
  \begin{tabular}{rc|rr|rr|rr|rr||rr|rr|rr|rr}
    \hline
        ~ &  ~ & \multicolumn{8}{c||}{\textbf{Unthresholded}} & \multicolumn{8}{c}{\textbf{Thresholded}} \\ \hline
        \rowcolor{graycolour!50}
         ~ & ~ & \multicolumn{2}{c|}{\textbf{70M}} & \multicolumn{2}{c|}{\textbf{125M}} & \multicolumn{2}{c|}{\textbf{250M}} & \multicolumn{2}{c||}{\textbf{500M}} & \multicolumn{2}{c|}{\textbf{70M}} & \multicolumn{2}{c|}{\textbf{125M}} & \multicolumn{2}{c|}{\textbf{250M}} & \multicolumn{2}{c}{\textbf{500M}} \\ 
        \rowcolor{graycolour!50}
        \textbf{Domain-Lang} & \textbf{Lang} & \textbf{D-L} & \multicolumn{1}{c|}{\textbf{L}} & \textbf{D-L} & \multicolumn{1}{c|}{\textbf{L}} & \textbf{D-L} & \multicolumn{1}{c|}{\textbf{L}} & \textbf{D-L} & \multicolumn{1}{c||}{\textbf{L}} & \textbf{D-L} & \multicolumn{1}{c|}{\textbf{L}} & \textbf{D-L} & \multicolumn{1}{c|}{\textbf{L}} & \textbf{D-L} & \multicolumn{1}{c|}{\textbf{L}} & \textbf{D-L} & \multicolumn{1}{c}{\textbf{L}} \\ \hline
        \makecell[{{r}}]{OSCAR-EN\\WIKI-EN} & EN & \makecell[{{r}}]{6.37\\7.61} & 13.98 & \makecell[{{r}}]{5.66\\8.30} & 13.96 & 
        \makecell[{{r}}]{1.41\\12.53} & 13.94 & \makecell[{{r}}]{1.82\\10.92} & 12.74 & \makecell[{{r}}]{7.54\\7.27} & 14.81 &\makecell[{{r}}]{6.84\\7.55} & 14.39 & \makecell[{{r}}]{6.29\\7.65} & 13.94 & \makecell[{{r}}]{6.41\\7.64} & 14.05\\ 
        \rowcolor{graycolour!50}
        \makecell[{{r}}]{OSCAR-ES\\WIKI-ES} & 
        ES & \makecell[{{r}}]{7.71\\5.80} & 13.51 & \makecell[{{r}}]{1.82\\8.13} & 9.95 & \makecell[{{r}}]{0.42\\8.58} & 9.00 & \makecell[{{r}}]{0.39\\13.14} & 13.53 & \makecell[{{r}}]{8.31\\6.43} & 14.74 & \makecell[{{r}}]{7.96\\7.21} & 15.17 & \makecell[{{r}}]{7.37\\9.91} & 17.28 & \makecell[{{r}}]{7.95\\9.88} & 17.83 \\
\makecell[{{r}}]{OSCAR-CA\\WIKI-CA} & CA & \makecell[{{r}}]{13.01\\6.84} & 19.85 & \makecell[{{r}}]{5.30\\6.16} & 11.46 & \makecell[{{r}}]{2.59\\1.80} & 4.39 & \makecell[{{r}}]{1.84\\4.85} & 6.69 & \makecell[{{r}}]{11.46\\7.62} & 19.08 & \makecell[{{r}}]{10.62\\7.55} & 18.17 & \makecell[{{r}}]{8.13\\8.14} & 16.27 & \makecell[{{r}}]{9.57\\9.08} & 18.65 \\
\rowcolor{graycolour!50}
\makecell[{{r}}]{OSCAR-GL\\WIKI-GL} & GL & \makecell[{{r}}]{0.99\\13.25} & 14.24 & \makecell[{{r}}]{0.72\\9.50} & 10.22 & \makecell[{{r}}]{0.78\\8.03} & 8.81 & \makecell[{{r}}]{0.62\\7.07} & 7.69 & \makecell[{{r}}]{6.37\\8.13} & 14.50 & \makecell[{{r}}]{6.65\\9.26} & 15.91 & \makecell[{{r}}]{7.30\\9.56} & 16.86 & \makecell[{{r}}]{7.56\\9.88} & 17.44 \\
\makecell[{{r}}]{OSCAR-EU\\WIKI-EU} & EU & \makecell[{{r}}]{9.50\\15.04} & 24.54 & \makecell[{{r}}]{7.85\\35.23} & 43.08 & \makecell[{{r}}]{7.30\\42.70} & 50.00 & \makecell[{{r}}]{0.28\\44.24} & 44.52 & \makecell[{{r}}]{11.25\\10.60} & 21.85 & \makecell[{{r}}]{9.75\\11.79} & 21.54 & \makecell[{{r}}]{7.22\\12.18} & 19.40 & \makecell[{{r}}]{6.81\\9.55} & 16.36 \\
\rowcolor{graycolour!50}
\makecell[r]{OSCAR-PT\\WIKI-PT} & PT & \makecell[{{r}}]{9.61\\4.27} & 13.88 & \makecell[{{r}}]{2.85\\8.48} & 11.33 & \makecell[{{r}}]{0.68\\13.18} & 13.86 & \makecell[{{r}}]{1.14\\13.70} & 14.84 & \makecell[{{r}}]{9.21\\5.83} & 15.04 & \makecell[{{r}}]{7.72\\7.14} & 14.86 & \makecell[{{r}}]{7.30\\8.95} & 16.25 & \makecell[{{r}}]{7.28\\8.39} & 15.67 \\ \hline
\multicolumn{2}{r|}{\textbf{Total}} & 100.0 & 100.0 & 100.0 & 100.0 & 100.0 & 100.0 & 100.0 & 100.0 & 100.0 & 100.0 & 100.0 & 100.0 & 100.0 & 100.0 & 100.0 & 100.0
\\ \hline
\rowcolor{graycolour!50}
        \multicolumn{2}{r|}{\textbf{$D_{KL}$ (D-L, base 500M)}} & 92.85 & - & 29.74 & - & 19.48 & - & 0.00 & - & 3.30 & - & 1.60 & - & 0.56 & - & 0.00 & - \\
        \rowcolor{graycolour!50}
        \multicolumn{2}{r|}{\textbf{$D_{KL}$ (L, base 500M)}} & - & 16.11 & - & 2.83 & - & 1.80 & - & 0.00 & - & 1.42 & - & 1.05 & - & 0.45 & - & 0.00 \\\hline
  \end{tabular}
  }
  \label{tab:threshold}
\end{table*}

\paragraph{Weight dynamics visualization}
To visually validate the weight stability achieved through our thresholded approach, Fig.~\ref{fig: unth vs th} compares the training dynamics of language-domain weights between our implementation and the original DoGE configuration. The side-by-side comparison demonstrates how thresholding prevents extreme divergences while maintaining the competition between sources and stable weight evolution. Notably, even if the optimization results in the largest weight for a high-resource language, the remaining are represented more fairly than with a natural ad hoc distribution. For example, the final weights of 13\% for Galician and 20\% for English in Fig.~\ref{fig: unth vs th}, in fact, correspond to an increase from a natural 0.19\%, and a decrease from a natural 44.65\%.

\begin{figure}[ht!]
    \centering\includegraphics[width=1.0\linewidth]{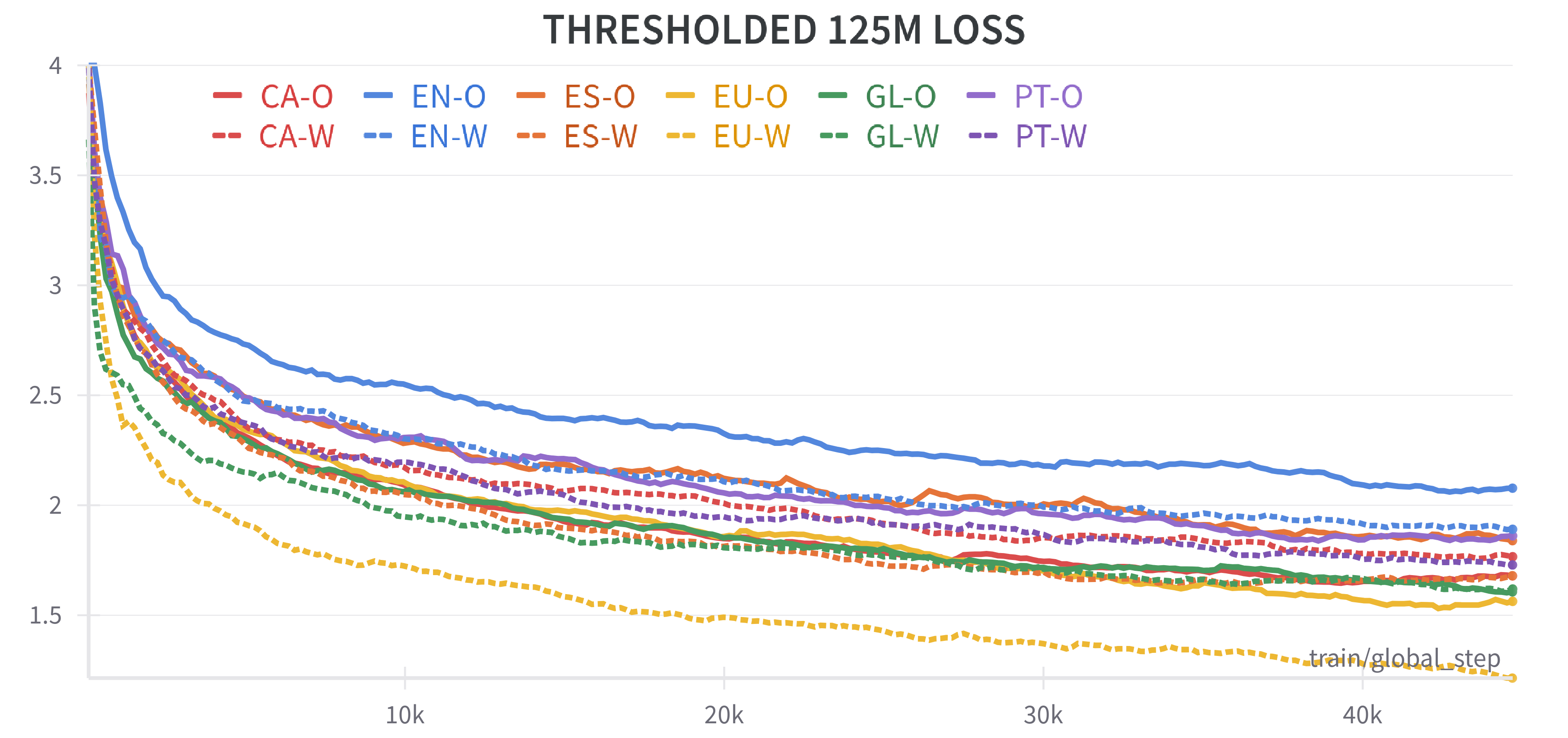}
    
    \caption{Thresholded proxy train loss per source. The dotted lines -- Wikipedia, continuous --  OSCAR.}
    \label{fig:loss_decrease}
\end{figure}

\begin{figure}[!ht]
 \centering
 \includegraphics[width=0.99\linewidth]{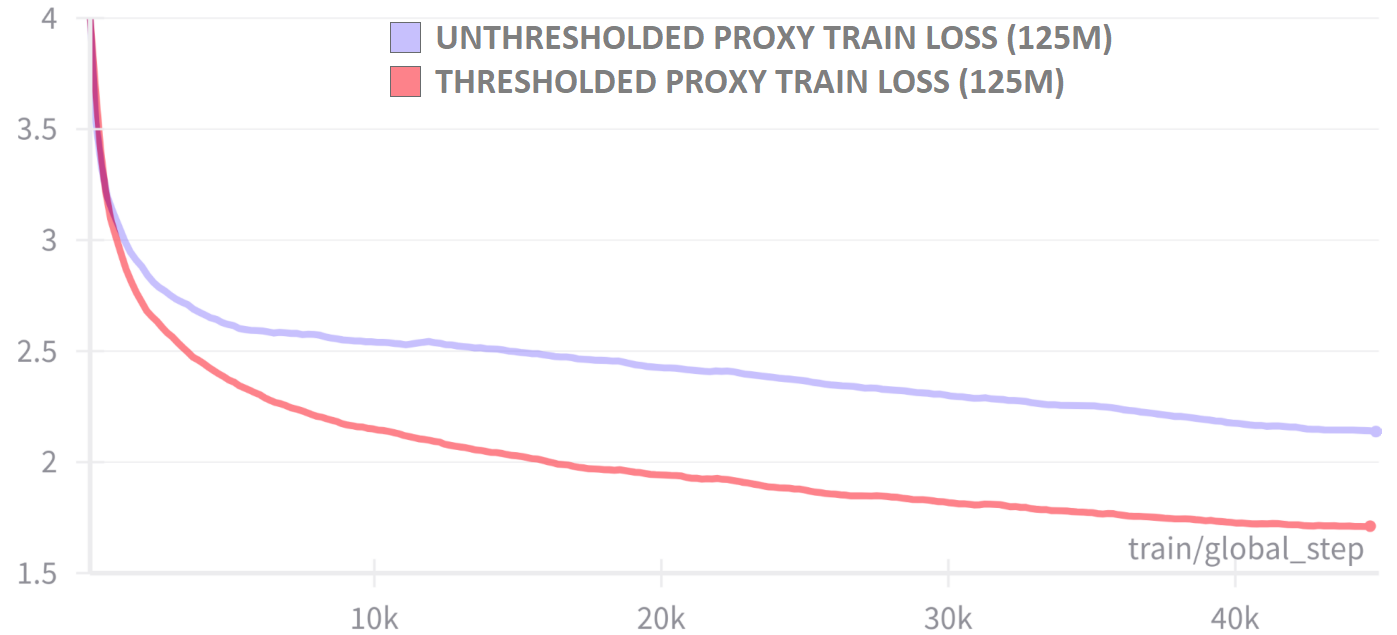}
    \caption{125M proxy train loss for DoGE and XDoGE.}
    \label{fig: train loss unth vs th}
\end{figure}

\paragraph{Training stability}
A critical requirement for any weighting adaptation is maintaining stable training dynamics. Fig.~\ref{fig:loss_decrease} provides this validation by showing the loss trajectories across all language sources in our thresholded setup. The coordinated downward trend confirms that our modifications preserve the fundamental training behavior: no language is destabilized. Fig.~\ref{fig: train loss unth vs th} quantifies how thresholding improves overall optimization.

\begin{figure}[!ht]
    \centering
\includegraphics[width=\columnwidth]{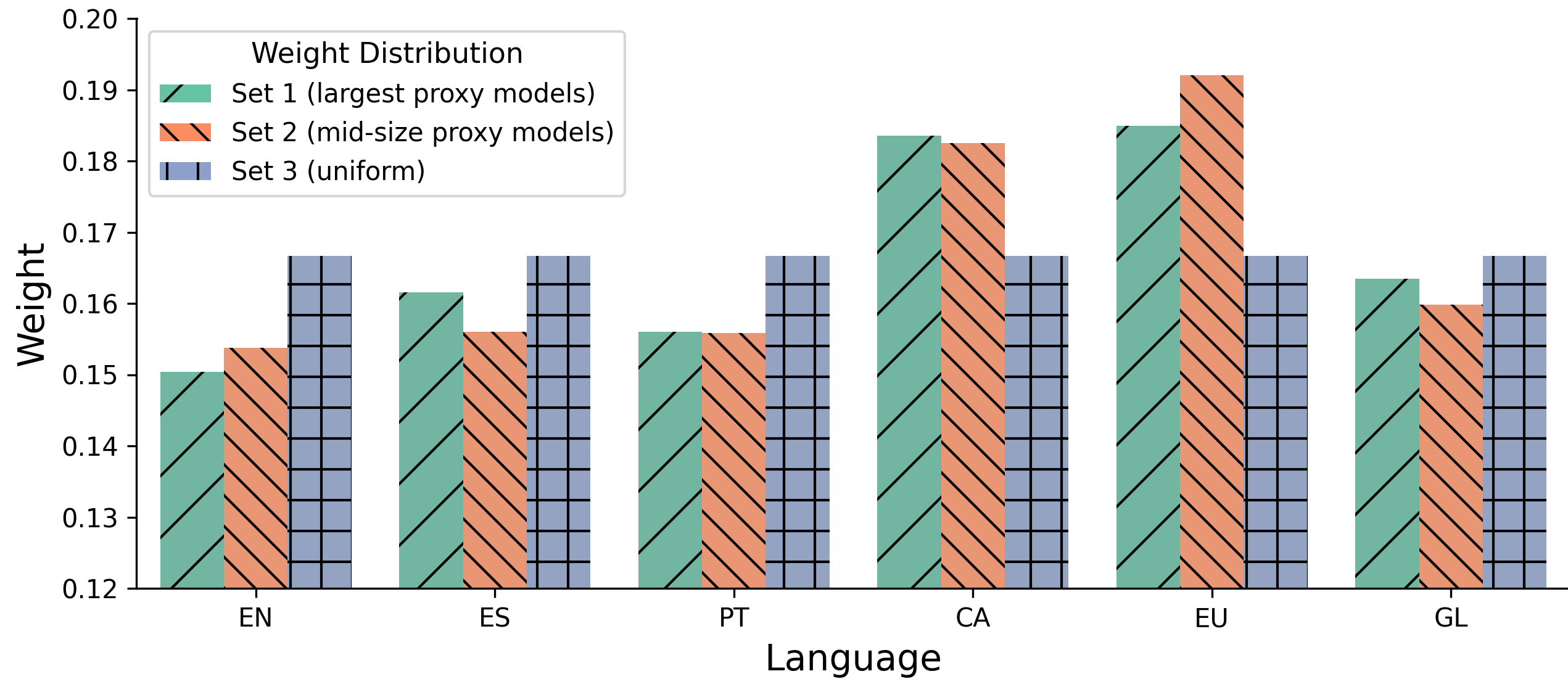}
    \caption{Selected weight distributions.}
    \label{fig:selected_weights}
\end{figure}

\paragraph{Model scaling and weight selection} 
We also observe in Table~\ref{tab:threshold} that \textit{\textbf{the larger the proxy model the more similar the distributions are and especially stable for the thresholded setup}}. Although intuitively they could be considered the best found, we opt for averaging across various proxy model sizes and different seed runs following \cite{Fan}. Since XDoGE features less stable convergence than DoGE across model sizes, we select two averaging schemes to test how sensitive the training is to the deviations in the language balancing: averaging over our four largest models (two 125M models of different seeds, 250M, and 500M models) and three middle-size models (500M model is excluded) using the last checkpoints.\footnote{Due to computing limitations we trained the 500M model only to up to 11k steps and smaller models to up to 50K steps. The number of models is also subject to available computing.} We do not include 70M models that give the largest weight dispersion with a distribution significantly diverging from those of larger models. We also consider a uniform distribution for baselines (i.e., sampling the same number of documents per language). The final sets of weights are provided in Fig.~\ref{fig:selected_weights}.

\begin{table}[ht!]
  \centering
  \caption{Perplexity scores (the lower the better) and IberoBench average accuracy scores (the higher the better) for the medium 500M base models.}
  \footnotesize
  \setlength{\tabcolsep}{4pt} 
  \resizebox{\columnwidth}{!}{
    \begin{tabular}{l|rrr|rrr}
      \hline
       & \multicolumn{3}{c}{\textbf{Perplexity}} & \multicolumn{3}{c}{\textbf{IberoBench (acc)}} \\
      \textbf{Language} & Bsl. & Set1 & Set2 & Bsl. & Set1 & Set2 \\
      \hline
      CA 
      & 14.43 & \textbf{14.42} & \textbf{14.42} & 41.01 & \textbf{41.21} & 40.29
      \\
      \hline
      PT
      & \textbf{31.33} & 31.52 & 32.30 & 51.73 & \textbf{53.19} & 43.66\\
      \hline
      EN
      & 31.56 & \textbf{31.40} & 32.13 & 38.15 & 38.03 & \textbf{38.87}\\
      \hline
      GL
      & \textbf{29.38} & 32.07 & 29.71 & 32.52 & \textbf{33.00} & 32.60\\
      \hline
      ES 
      & 35.15 & 35.30 & \textbf{35.13} & 43.43 & 44.17 & \textbf{44.73}\\
      \hline
      EU 
      & \textbf{23.11} & 23.61 & 23.55 & 37.06 & 36.12 & \textbf{37.24}\\
      \hline
      \hline
      \textbf{Total Avg} & 
      \textbf{27.49} & 28.05 & \textit{27.87} & \textit{40.65} & \textbf{40.95} & 39.57\\
      \hline
    \end{tabular}
  }
  \label{tab:perplexity_iberobench_base_final}
\end{table}

\begin{table}[ht!]
  \centering
  \caption{IberoBench average accuracy scores (the higher the better) for small 250M and large 900M base models.}
  \footnotesize
  \setlength{\tabcolsep}{4pt} 
  \resizebox{\columnwidth}{!}{
    \begin{tabular}{l|rrr|rrr}
      \hline
       & \multicolumn{3}{c}{\textbf{250M}} & 
       \multicolumn{3}{c}{\textbf{900M}} \\
       \textbf{Language} & Bsl. & Set1 & Set2 & Bsl. & Set1 & Set2 \\
      \hline
      CA 
      & \textbf{39.88} & 39.32 & 39.40 & 41.85 & 40.88 & \textbf{42.07}
      \\
      \hline
      PT
      & 46.00 & \textbf{52.32} & 48.85 & 48.64 & \textbf{48.83} & 47.24\\
      \hline
      EN
      & 37.03 & \textbf{37.18} & 36.25 & 38.17 & \textbf{38.65} & 38.22\\
      \hline
      GL
      & 32.64 & 33.65 & \textbf{33.68} & 31.05 & \textbf{32.10} & 31.22\\
      \hline
      ES 
      & \textbf{45.02} & 43.63 & 43.89 & \textbf{43.15} & 40.64 & 41.87\\
      \hline
      EU 
      & 36.75 & \textbf{36.87} & 36.24 & \textbf{37.07} & 36.47 & 36.76\\
      \hline
      \hline
      \textbf{Total Avg} & 
      39.55 & \textbf{40.50} & \textit{39.72} & \textbf{39.99} & \textit{39.60} & 39.56\\
      \hline
    \end{tabular}
  }
  \label{tab:iberobench_bases_final}
\end{table}

\subsection{Base Models}

We trained base models of varying sizes using three weight setups from Fig.~\ref{fig:selected_weights}. Tables~\ref{tab:perplexity_iberobench_base_final}~and~\ref{tab:iberobench_bases_final} compare Set1 and Set2 against the uniform-set baseline.\footnote{The evaluations are performed on all accuracy tasks on IberoBench.}

500M XDoGE exhibits the highest accuracy across all languages, but shows lower perplexity for some (whereas 250M surpasses the baseline across all tasks). We do not find stable language-set associations across model sizes, but, in general, more languages benefit from Set1.

\begin{figure*}[ht!]
  \centering
  \includegraphics[width=0.37\linewidth]{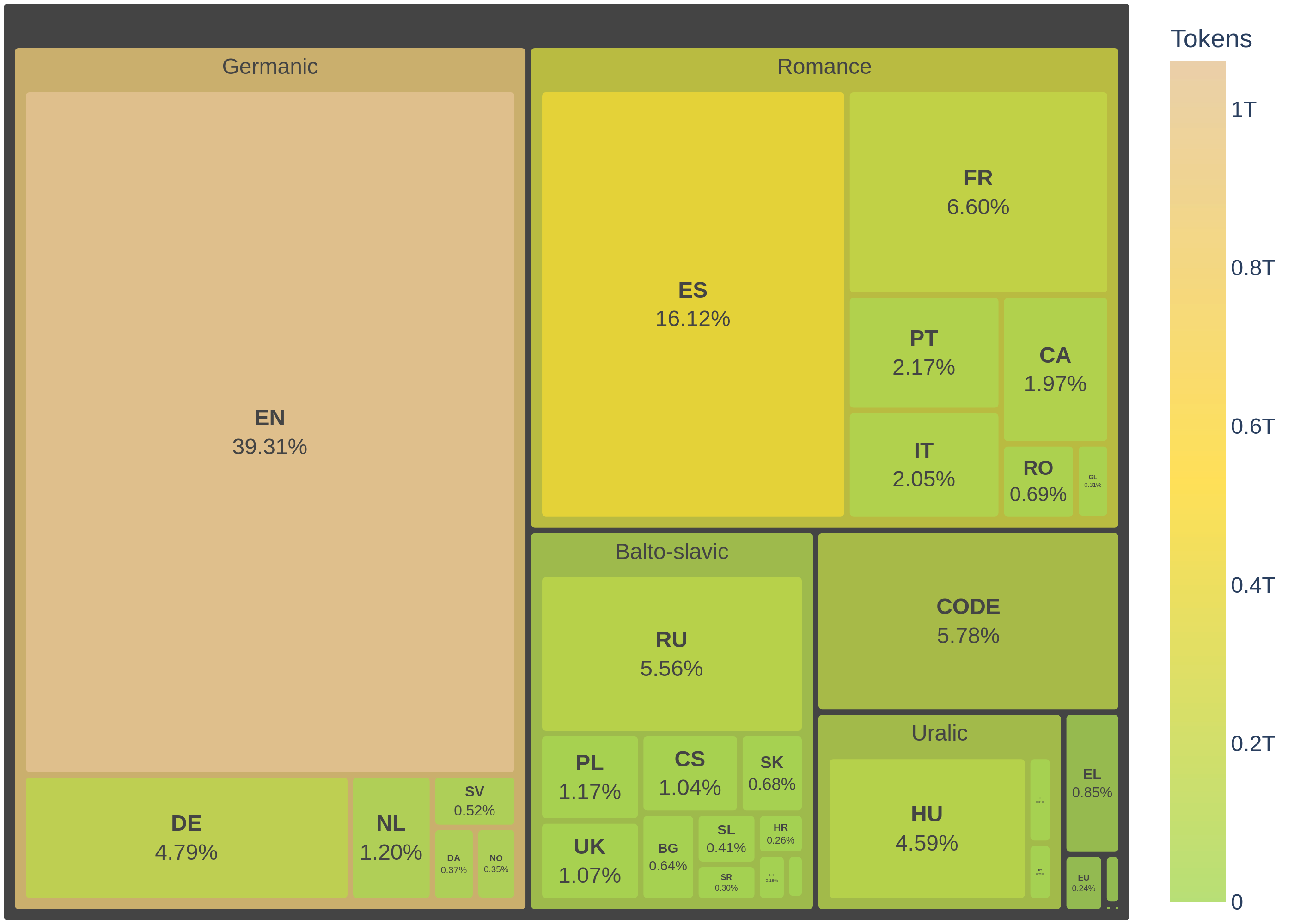}
  \includegraphics[width=0.58\linewidth]{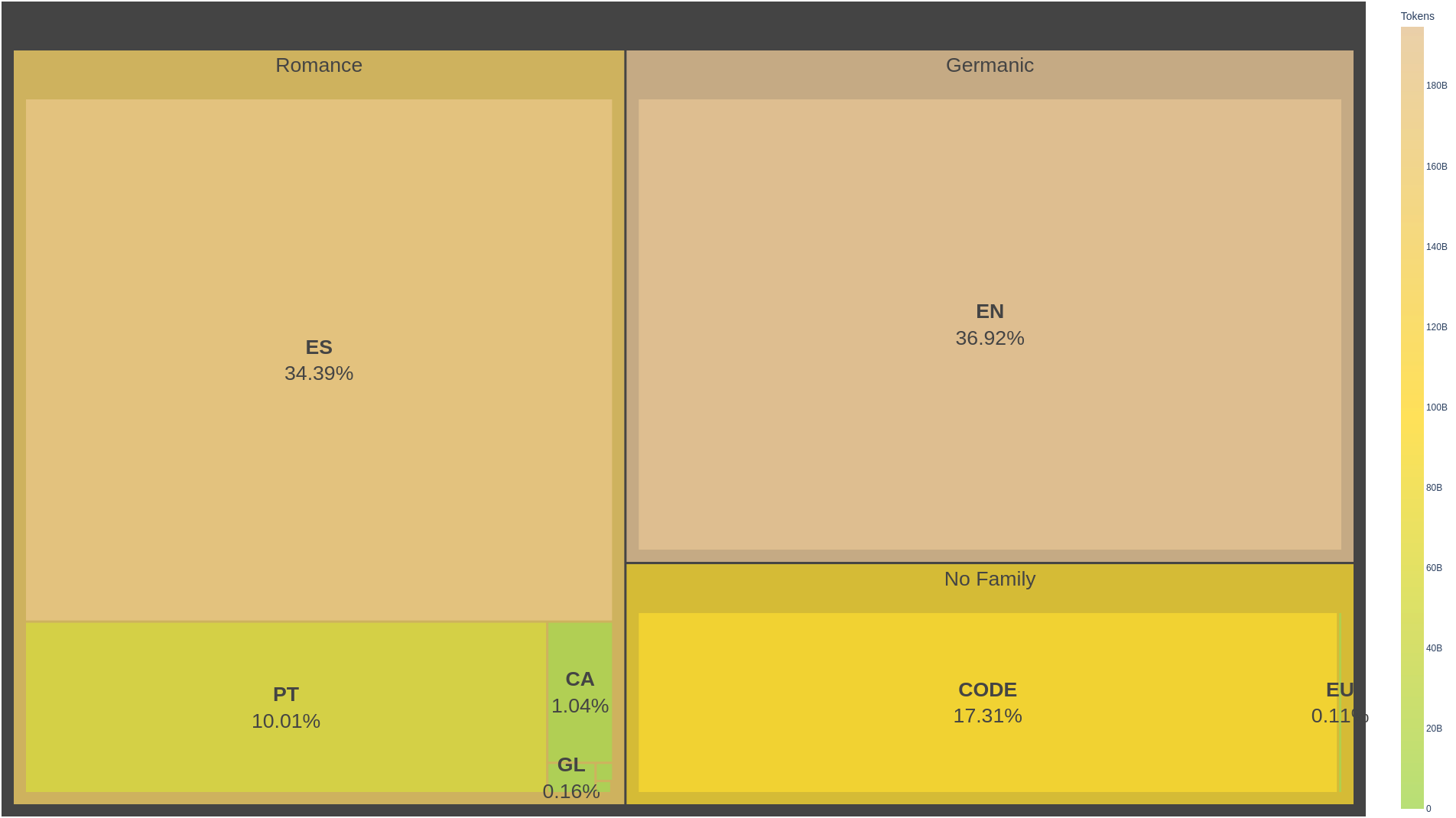}
  \caption{Data distribution weights for each language during Salamandra pre-training (left) and 7b model pre-training (right). The total number of tokens for each model are \(2.68T\) and \(542B\), respectively.}
  \label{fig:pata negra data dist}
\end{figure*}

\subsection{IberianLLM-7b Pre-training} \label{sec:iberianllm_pretraining}

This subsection outlines the architecture, technical setup, and training methodology employed to develop IberianLLM. It also discusses the rationale behind our design choices, particularly in handling the unique challenges posed by low-resource Iberian languages.

\paragraph{Challenges in Training LLMs with Low-Resource Languages}

Training large language models for low-resource languages is inherently challenging due to the limited availability of high-quality data. 

Maximizing data utilization is crucial, yet even with careful curation, the overall volume for low-resource languages remains much smaller compared to high-resource ones like Spanish or English. A common strategy to mitigate this issue is to pre-train models on diverse linguistic distributions while filtering out low-quality sources \cite{Micallef}.

Table~\ref{tab:data} shows that the token counts for each language clearly indicate the impracticality of training a very large model from scratch with our available data. In particular, when aiming for a balanced representation across languages, one must consider that effective large-scale training typically requires around 30 tokens per parameter \cite{Hoffmann}. Given that Galician, the lowest-resourced language in our set, comprises approximately 213 million tokens, a balanced approach would cap each language to a similar token count. This results in a combined dataset of roughly 1.28 billion tokens across six languages, for a total token consumption of about 3.84 billion tokens over three epochs. According to the heuristic, this corresponds to a model with approximately 128 million parameters, ensuring that no language is oversampled at the expense of another.

\paragraph{Optimizing Training Data for Target Languages}

The Salamandra family of models was pre-trained on data spanning 35 European languages and code, which includes all of our target languages. However, instead of adopting a highly multilingual model, we opted to focus on a more specialized set: Iberian languages, English (selected for its abundant data), and code. In addition to our primary targets -- English, Spanish, Catalan, Basque, Galician, and Portuguese -- we also included Occitan, Aragonese, and Balearic. These additional languages, due to their close linguistic ties to our main Iberian languages, help enhance the model's ability to capture related language structures.

Notably, while the Salamandra models upscaled Iberian languages by doubling their data share, our approach inherently centers on these languages, eliminating the need for further upscaling. Moreover, to avoid an English-centric bias, we adjusted the composition of the English data. Instead of merely subsampling the original English dataset used in Salamandra, we replaced the Oscar portion with the higher-quality FineWeb-Edu dataset \cite{Penedo}. This change reduced the overall quantity of English data -- bringing it closer in line with Spanish, the next most-resourced target language -- while significantly improving data quality.

Overall, the IberianLLM model was pre-trained on approximately \(0.53T\) unique tokens, in contrast to the \(2.6T\) unique tokens used for the Salamandra models. Fig.~\ref{fig:pata negra data dist} compares the language distribution in our pre-training data with that of the Salamandra family, underscoring our targeted focus.

\paragraph{Architecture \& Training Strategy}

\begin{table*}
  \centering
  \caption{\label{tab:base_dimensions} Dimensions of the base models and the large architectures we created and used for CPT experiments. These models have been created using the Salamandra architecture.}
  \begin{tabular}{llllll}
    \hline
    \textbf{Model ID} & \textbf{Base-250M} & \textbf{Base-500M} &
    \textbf{Base-900M} & \textbf{Salamandra-2b} & \textbf{IberianLLM-7b}\\
    \hline
    Parameters & 277,881,344 & 497,050,368 & 901,825,536 & 2,253,490,176 & 7,768,117,224\\
    Layers & 8 & 16 & 24 & 24 & 32 \\
    Hidden Size & 512 & 768 & 1,024 & 2,048 & 4,096 \\
    FFN Size & 768 & 2,048 & 4,096 & 5,440 & 11,008\\
    Attention Heads & 8 & 16 & 16 & 16 & 32\\
    K/V Heads & 4 & 8  & 8 & 16 & 8\\
    Context Length & 8,192 & 8,192 & 8,192 & 8,192 & 8,192\\
    Vocabulary Size & 256,000 & 256,000 & 256,000 & 256,000 & 256,000 \\
    \hline
  \end{tabular}
  
\end{table*}

Table~\ref{tab:base_dimensions} shows the architectural configuration of our model, which mirrors the 7b version of the Salamandra model.

We trained IberianLLM using the NeMo framework, following a multi-epoch strategy similar to that of the Salamandra models. Pre-training was conducted over three epochs to ensure comprehensive exposure to our curated dataset. Table~\ref{tab:pata_negra_training_config} outlines the key training parameters.

Further details on the tokenizer implementation and training can be found in Section~\ref{subsec: technical setup}, and additional insights are provided in the Salamandra technical report \cite{Gonzalez}.

\begin{table}[ht]
  \centering
  \caption{ 
  Training hyperparameters and optimizer settings for pre‑training the IberianLLM.}
  \begin{tabular}{ll}
    \hline
    \textbf{Training Configuration} & \textbf{Value} \\
    \hline
    Epochs and Steps & 3 training epochs, 377215 steps \\
    Context Length & 8192 \\
    Global Batch Size & 512 \\
    Optimizer Name & Distributed Fused Adam \\
    Learning Rate (lr) & \(3.0e^{-4}\) \\
    Weight Decay & 0.1 \\
    Betas & [0.9, 0.95] \\
    Scheduler & CosineAnnealing \\
    Warmup Steps & 2000 \\
    Minimum Learning Rate & \(3.0e^{-5}\) \\
    \hline
  \end{tabular}
  
  \label{tab:pata_negra_training_config}
\end{table}

\subsection{Continual Pretraining Phase}

Within CPT, we used only Set1 weights of XDoGE (since they showed superiority in most cases across various base model sizes). We focus on comparing with uniform weights and ad hoc weights (i.e., defined by the original data distribution).

\begin{figure}
    \centering
    \includegraphics[width=1.00\linewidth]{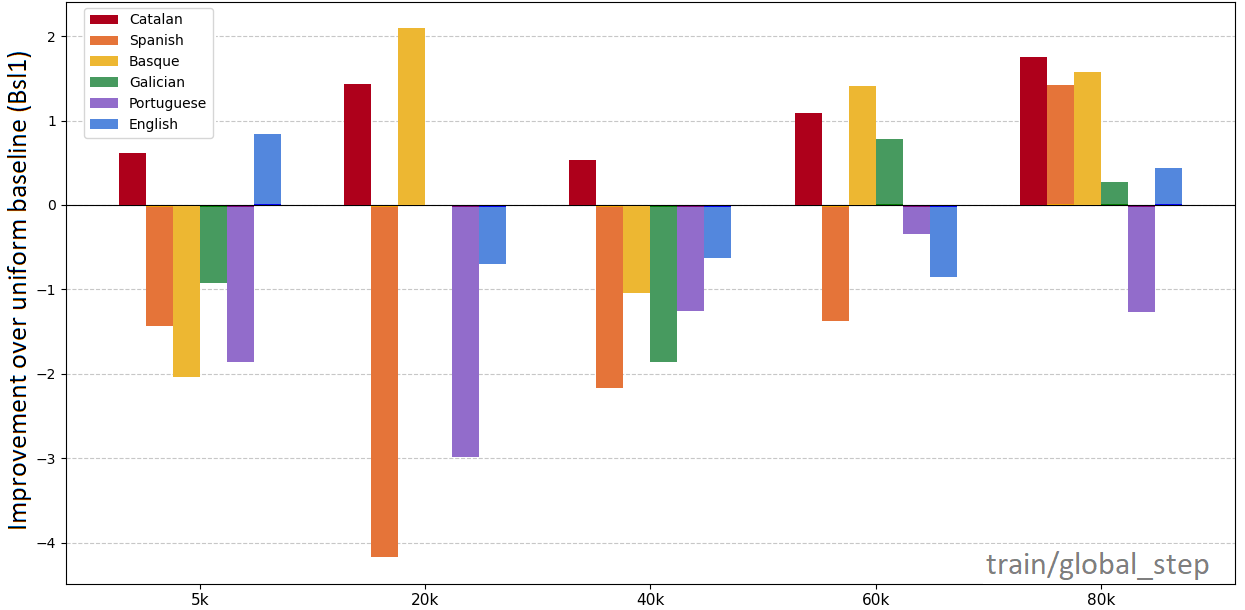}
    \caption{Gains (in \%) of the Salamandra-2b CPT with XDoGE-Set1 over Bsl1.}
    \label{fig:2B_cpt_improvement_in_percentages}
\end{figure}

Table~\ref{tab:perplexity_iberobench_cpt_final} summarizes the evaluation, which shows that Salamandra-2b with XDoGE achieves a lower or equal perplexity in all languages, suggesting improved overall generalization on the target languages. On IberoBench tasks, our CPTs achieve higher average scores over the original Salamandra-2b and IberianLLM-7b and the baselines (see also the relative improvements in dynamic over the uniform baseline in Fig.~\ref{fig:2B_cpt_improvement_in_percentages}), with Catalan, Spanish and Basque benefiting the most. English encounters under-sampling, while Portuguese and Galician received lower weights, preventing them from sufficiently reinforcing each other.

\begin{table}[ht]
  \centering
  \caption{Perplexity (lower is better) and IberoBench scores (higher is better) for each language between the original checkpoint from which the CPT was performed (No-CPT) and CPT with uniform (Bsl-1), ad hoc (Bsl-2), and XDoGE weights.}
  \footnotesize
  \setlength{\tabcolsep}{4pt} 
  \resizebox{\columnwidth}{!}{%
    \begin{tabular}{l|rr|rrr|rrr}
      \hline
      & \multicolumn{5}{c|}{\textbf{Salamandra-2B}} & \multicolumn{3}{c}{\textbf{IberianLLM-7B}} \\
      & \multicolumn{2}{c|}{\textbf{Perplexity}} & \multicolumn{3}{c|}{\textbf{IberoBench}} & \multicolumn{3}{c}{\textbf{IberoBench (acc)}} \\
      \textbf{Language} & Bsl-1 & XDoGE & No-CPT & Bsl-1 & XDoGE & No-CPT & Bsl-2 & XDoGE \\
      \hline
      CA 
      & 6.42      & \textbf{6.37}
      & 51.66 &  \textit{51.78} & \textbf{52.69} & \textit{53.76} & 53.40 & \textbf{54.77}
      \\
      \hline
      PT
      & 11.27     & \textbf{11.16}
      & \textbf{53.34} &  \textit{52.87} & 52.20 & 48.87 & \textbf{54.01} & \textit{52.94} \\
      \hline
      EN
      & \textbf{10.80} & \textbf{10.80}
      & \textbf{46.49} &  45.51 & \textit{45.71} & \textit{52.24} & \textbf{53.08} & 52.10 \\
      \hline
      GL
      & 7.79      & \textbf{7.66}
      & 29.54 &  \textit{29.66} & \textbf{29.74} & \textit{34.31} & \textbf{34.64} & 34.28\\
      \hline
      ES 
      & 12.19     & \textbf{12.15}
      & \textit{50.19} &  50.14 & \textbf{50.85} & 46.75 & \textit{48.70} & \textbf{50.46}\\
      \hline
      EU 
      & 5.21      & \textbf{5.20}
      & 38.47 &  \textit{39.35} & \textbf{39.97} & 39.76 & \textit{41.03} & \textbf{41.44}\\
      \hline
      \hline
      \textbf{Total Avg} & 
      8.95      & \textbf{8.89}
      & \textit{44.95} &  44.89 & \textbf{45.19}$^{\mathrm{a}}$ & 45.95 & \textit{47.48} & \textbf{47.67}$^{\mathrm{b}}$\\
      \hline
      \multicolumn{9}{l}{$^{\mathrm{a}}$The differences with No-CPT and Bsl-1 are statistically significant at the 5\% level.}\\
      \multicolumn{9}{l}{$^{\mathrm{b}}$The difference with No-CPT is statistically significant at the 5\% level.}
    \end{tabular}%
  }
  \label{tab:perplexity_iberobench_cpt_final}
\end{table}

As we used the updated web data for all CPT setups, including baselines, we conclude that the new information is not solely responsible for the gains over the original models (\textit{No-CPT}). This justifies the key role of our weights. Over-repetition of the data also contributes to the gains. Although uniform weights also imply this, we observe that our weights benefit the model in a longer training run, delaying model degradation: a slight decrease after 60k-80k steps for XDoGE and a significant decline after 40k-60k steps for uniform is seen for most languages.

\subsection{Evaluation and Analysis of IberianLLM}
\label{sec:pata_negra_analysis}

To position the IberianLLM-7b introduced in Section~\ref{subsec: technical setup}, we provide a summary of the IberoBench evaluations in Table~\ref{tab:pata-negra-summary}, comparing it with the Salamandra-7b.

\begin{table*}[ht]
  \centering
  \caption{Model comparison with relative improvements on IberoBench. Iber. = IberianLLM-7b, Sala. = Salamandra-7b. Positive percentages indicate when IberianLLM-7b overcomes Salamandra-7b. All values averaged per group.}
  \footnotesize
  \setlength{\tabcolsep}{4pt}
  \resizebox{\textwidth}{!}{
    \begin{tabular}{l|cc|cc|cc|cc|cc|cc|cc|cc|cc}
      \hline
      & \multicolumn{2}{c}{\textbf{Flores Iberian}} & \multicolumn{2}{c}{\textbf{Flores Non-Iber}} & \multicolumn{2}{c}{\textbf{Belebele}} & \multicolumn{2}{c}{\textbf{English}} & \multicolumn{2}{c}{\textbf{Catalan}} & \multicolumn{2}{c}{\textbf{Spanish}} & \multicolumn{2}{c}{\textbf{Basque}} & \multicolumn{2}{c}{\textbf{Galician}} & \multicolumn{2}{c}{\textbf{Portuguese}} \\
      & Iber. & Sala. & Iber. & Sala. & Iber. & Sala. & Iber. & Sala. & Iber. & Sala. & Iber. & Sala. & Iber. & Sala. & Iber. & Sala. & Iber. & Sala. \\
      \hline
      Scores & 
      23.63 & 25.42 & 
      18.42 & 24.23 & 
      28.04 & 25.91 & 
      43.84 & 43.46 & 
      50.99 & 52.70 & 
      35.37 & 36.03 & 
      37.36 & 39.87 & 
      26.76 & 27.12 &     
      58.80 & 63.78 \\ 
      \hline
      $\Delta\%$ & \multicolumn{2}{c}{\textbf{-7.04\%}} & \multicolumn{2}{c}{\textbf{-23.98\%}} & \multicolumn{2}{c}{\textbf{+8.22\%}} & \multicolumn{2}{c}{\textbf{+0.87\%}} & \multicolumn{2}{c}{\textbf{-3.24\%}} & \multicolumn{2}{c}{\textbf{-3.44\%}} & \multicolumn{2}{c}{\textbf{-6.29\%}} & \multicolumn{2}{c}{\textbf{-1.33\%}} & \multicolumn{2}{c}{\textbf{-7.81\%}} \\
      \hline
    \end{tabular}
  }
  \label{tab:pata-negra-summary}
\end{table*}

Our key findings are the following:
(i) IberianLLM notably outperforms Salamandra in Belebele multilingual reading comprehension;
(ii) trained with only 0.53T unique tokens (vs. Salamandra's 2.6T), IberianLLM achieves 95.58\% of Salamandra's average score on Iberian tasks. This suggests that our targeted training yields roughly 4.9× greater token efficiency; (iii) IberianLLM shows competence in few-shot translation involving never-seen languages like German and French, achieving 76\% of Salamandra's performance on average despite the presumable absence of these languages and parallel data in the training corpus. This highlights the model's strong cross-lingual capabilities.

\section{Discussion}

The results presented in Table \ref{tab:perplexity_iberobench_cpt_final} and illustrated in Fig.~\ref{fig:2B_cpt_improvement_in_percentages} highlight the success of our measures to enhance linguistic inclusivity in LLMs. With two differently sized models capable of supporting various numbers of languages, we verified that they can be improved in more than half of the target languages. This aligns with the original DoGE's ambition to excel in more than half of the target domains \cite{Fan}. Continual pretraining with XDoGE weights outperforms a uniform baseline (Bsl-1) that neglects language transferability, across five of six target languages (with a statistical significance at the 5\% level). It is also not inferior to training with ad hoc (natural) weights (Bsl-2), on average across languages, while the advantage is the ability to steer the model towards highly weighted languages (Catalan and Basque in our case) and towards languages sharing commonalities (Spanish). Thus, if a specific combination of target languages makes proxy models converge to multiple options, we recommend choosing one that best satisfies needs in specific downstream use cases. We presume it is unlikely to excel in all languages at once by adjusting language sampling weights alone.

The results also offer several insights validating our design choices.

Firstly, our thresholded DoGE implementation reduces weight variance, which stabilizes proxy training and substantially decreases the loss compared to the original approach (Fig.~\ref{fig: train loss unth vs th}). On the contrary, unconstrained proxy training leads to considerable discrepancies in weights, making them far unsuitable for real-world data volumes.

Secondly, applying the XDoGE framework to \textit{pre-trained} models yields lower perplexity and higher scores on downstream tasks more consistently than with \textit{base} models (cf. Tables \ref{tab:perplexity_iberobench_base_final}, \ref{tab:iberobench_bases_final} and \ref{tab:perplexity_iberobench_cpt_final}), suggesting its better suitability for continual training, especially by delaying model degradation caused by data over-repetition.

Finally, the full-size training is not very sensitive to slight changes in weights overall (cf. scores for Set1 vs Set2 in Table~\ref{tab:perplexity_iberobench_base_final}). The exclusion in our case was for Portuguese, with a 10-point drop on downstream tasks. We assume this happened due to a simultaneous decrease in the weights of the supporting Spanish and Galician languages.

\section{Conclusion and Future Work}

We introduced XDoGE -- an extension of the DoGE algorithm for multilingual pre-training that finds the optimized set of language weights that enhances cross-lingual transfer capability. In line with previous work, applying learnt weights for full-size training from scratch yields gains for only about half of the targets. Our proposal to use the weights in continual pre-training appears more beneficial, resulting in lower validation perplexity and improved downstream performance compared to the baselines for the majority of targets. We result in a series of LLMs of different sizes and origins, centred on Iberian languages and English.

For future work, we plan to explore adapting highly multilingual models to target languages while compromising less on non-target languages. Specifically, we will compare holistic language inclusion with a staged approach to changing target languages. We will also consider explicitly using language-family correspondences to improve the capture of interdependencies among languages.

\section{Limitations}
\label{app:limitations}

While our approach demonstrates the potential of optimizing multilingual data distributions in a principled manner, there are several limitations that we consider when interpreting the findings and designing future research.

\paragraph{Data availability as a contributing factor for weight estimation} The XDoGE framework is originally meant to be efficient when estimating the weights with the proxy models, making the scale of the proxy models too small to see all available data. This is particularly relevant for low-resource languages, where data repetition is often necessary for effective pre-training. Since proxy models do not necessarily replicate the availability of data across languages, they may not fully capture the impact of data scarcity and repetition.
While excluding data availability as a factor in estimating the language weights allows for clearer language interaction modeling, it does not address practical constraints in large-scale multilingual pre-training. Although our experiments have shown that controlled repetition can enhance language model learning in low-resource languages, future research should explore the impact of integrating data availability into the weight computation process. 
In particular, when computing weights for continual pre-training (CPT) experiments, it may be beneficial to train proxy models starting with the same weights used during pre-training instead of initializing them with uniform weights, as in our current setup. This could enable more accurate estimation of the optimal data mixture for CPT experiments.

\paragraph{Better efficiency in highly multilingual scenarios} The computational cost of estimating \(W\) increases with both model size and the number of data sources. While the original DoGE framework was applied to only 7 sources for domain-weight estimation in a monolingual setting, we scaled this to 12 domains across 6 languages. However, extending this approach to highly multilingual language models remains a challenge. Future research with the XDoGE or similar frameworks should take into account the scalability of data size and sources while maintaining their benefits for generalization.

\paragraph{Algorithm overhead clarity} The optimization procedure that our method implies may not feel fully justified compared to the baselines that are straightforward to apply. Still, we must consider that we work with models with only up to 7 billion parameters: it is challenging to develop advanced language capabilities, such as reasoning, needed for downstream tasks; therefore, the improvement we observe is limited. We expect that scaling the conclusions to larger models would considerably save computational resources. In this case, the XDoGE proxy training time becomes truly negligible, clarifying the contribution of experimenting with smaller models.

\section{Acknowledgements}

This work was funded by the Ministerio para la Transformación Digital y de la Función Pública and Plan de Recuperación, Transformación y Resiliencia - Funded by EU – NextGenerationEU within the framework of the project Modelos del Lenguaje and the ILENIA Project with reference 2022/TL22/00215337, 2022/TL22/00215336, 2022/TL22/00215335, 2022/TL22/00215334. It was also funded by the Project Desarrollo de Modelos ALIA with the framework of the Plan Nacional de Tecnologías de Lenguaje -ENIA 2024 and PRTR, NextGeneration EU, Resol. SEDIA 19.08.2024.
We also acknowledge the EuroHPC Joint Undertaking and the Spanish Supercomputing Network for awarding us access to MareNostrum5 and the technical support provided by the Barcelona Supercomputing Center (EHPC-EXT-2024E01-009, EHPC-AI-2024A05-048, RES-IM-2024-2-0031, RES-IM-2024-3-0021).

\vspace{12pt}

\end{document}